\pdfoutput=1

\documentclass[11pt]{article}
\usepackage[final]{acl}

\usepackage{times}
\usepackage{latexsym}
\usepackage{graphicx}
\usepackage{cleveref}
\usepackage{hyperref}
\usepackage[T1]{fontenc}
\usepackage{tcolorbox}
\usepackage{color, colortbl}
\usepackage{amssymb}

\usepackage{array}
\usepackage{makecell}

\usepackage[utf8]{inputenc}

\usepackage{microtype}

\usepackage{inconsolata}
\usepackage{booktabs} 
\usepackage{multirow}
\usepackage{enumitem}
\usepackage{subcaption}

\usepackage{fancyvrb}
\usepackage{fvextra}
\usepackage{amssymb}
\usepackage{pifont}
\usepackage{xstring}
\usepackage{xspace}
\newcommand{\cmark}{\ding{51}}%
\newcommand{\xmark}{\ding{55}}%

\newcommand{\github}[1]{%
  \StrBehind{#1}{https://github.com/}[\myGitHubPathTemp]%
  \mbox{%
    \xspace
    \raisebox{-0.6ex}{\includegraphics[width=1.2em, height=1.2em]{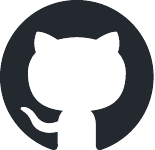}}%
    \hspace{0.4em}%
    \href{#1}{\texttt{\nolinkurl{\myGitHubPathTemp}}}%
  }%
}

\newcommand{\fngithub}[1]{%
  \StrBehind{#1}{https://github.com/}[\myGitHubPathTemp]%
  \mbox{%
    \xspace
    \raisebox{-0.4ex}{\includegraphics[width=1em, height=1em]{figures/github.pdf}}%
    \hspace{0.2em}%
    \href{#1}{\footnotesize\texttt{\nolinkurl{\myGitHubPathTemp}}}%
  }%
}

\newcommand{\huggingface}[1]{%
  \StrBehind{#1}{https://huggingface.co/}[\myHfPathTemp]%
  \xspace
  \hspace{-0.4em}
  \mbox{\raisebox{-0.8ex}{\includegraphics[width=1.35em, height=1.35em]{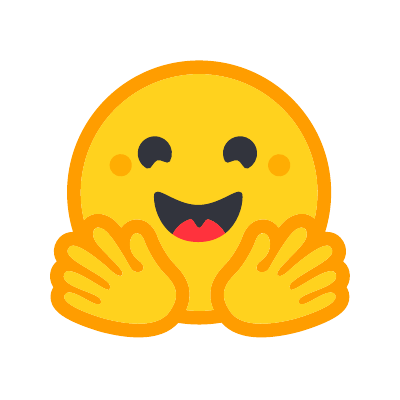}}\hspace{0.1em}}%
  \href{#1}{\texttt{\nolinkurl{\myHfPathTemp}}}%
}

\newcommand{\ku}{\raisebox{-.2ex}{\includegraphics[height=0.8em]{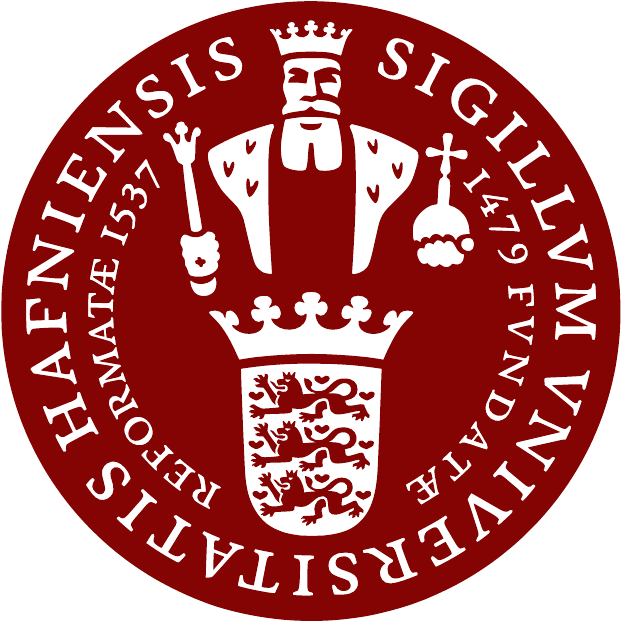}}}
\newcommand{\berlin}{\raisebox{-.2ex}{\includegraphics[height=0.6em]{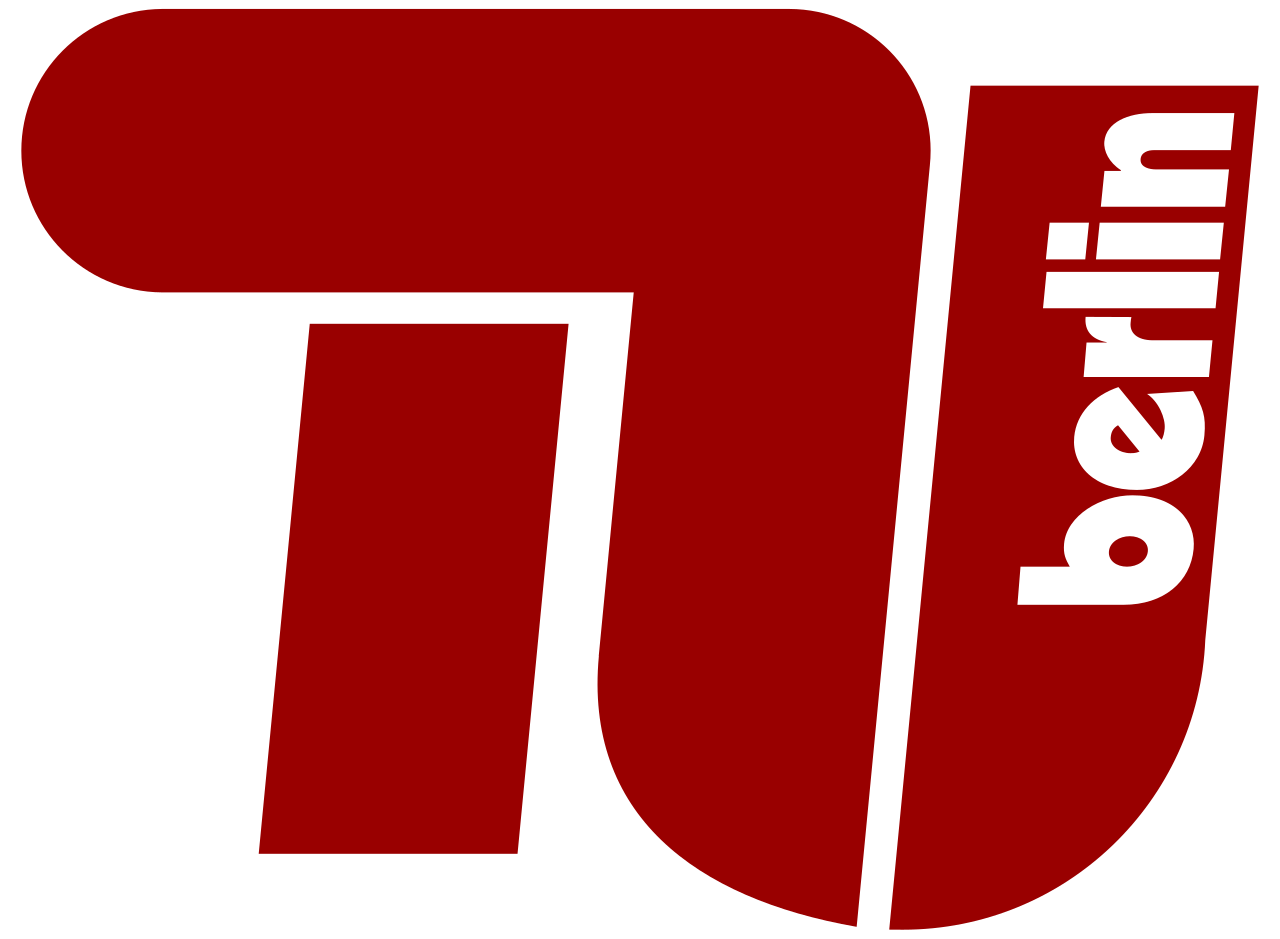}}}
\newcommand{\mitlogo}{\raisebox{-.2ex}{\includegraphics[height=0.6em]{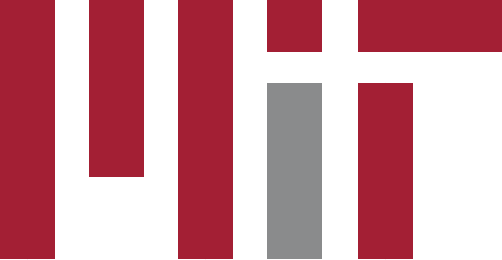}}}
\newcommand{\alephalph}{\raisebox{-.2ex}{\includegraphics[height=0.6em]{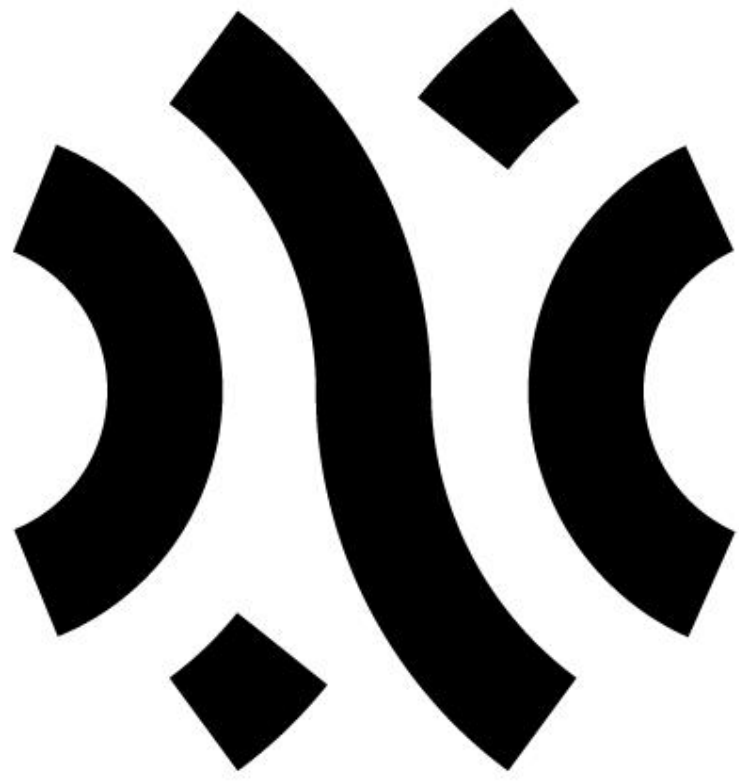}}}
\newcommand{\circa}{{\raise.17ex\hbox{$\scriptstyle\sim$}}}

\title{Trick or Neat:\\Adversarial Ambiguity and Language Model Evaluation}

\author{Antonia Karamolegkou$^{\ku}$ \ \ Oliver Eberle$^{\berlin}$ \ \ Phillip Rust$^{\ku}$ \ \ Carina Kauf $^{\mitlogo}$ $^{\alephalph}$ \ \ Anders S{\o}gaard$^{\ku}$ \vspace{0.2cm} \\
        $^{\ku}$University of Copenhagen \ \ 
        $^{\berlin}$Technische Universit\"{a}t Berlin \\
        $^{\mitlogo}$Massachusetts Institute of Technology 
        $^{\alephalph}$Aleph Alpha Research\\ 
        \small Correspondence: \href{mailto:antka@di.ku.dk}{\texttt{antka@di.ku.dk}}
}

\begin{document}
\maketitle
\begin{abstract}

Detecting ambiguity is important for language understanding, including uncertainty estimation, humour detection, and processing garden path sentences. We assess language models' sensitivity to ambiguity by introducing an adversarial ambiguity dataset that includes syntactic, lexical, and phonological ambiguities along with adversarial variations (e.g., word-order changes, synonym replacements, and random-based alterations). Our findings show that direct prompting fails to robustly identify ambiguity, while linear probes trained on model representations can decode ambiguity with high accuracy, sometimes exceeding 90\%. Our results offer insights into the prompting paradigm and how language models encode ambiguity at different layers. We release both our code and data: \fngithub{https://github.com/coastalcph/lm_ambiguity}.
\end{abstract}

\section{Introduction} \label{sec:intro}

Linguistic utterances often have ambiguous meanings, but our world knowledge helps us resolve them. Consider the ambiguous sentence in~\ref{instr}, and the unambiguous alterations in~\ref{NPmodifier} and~\ref{highAttach}:

\begin{enumerate}[label=(\arabic*)]\label{examples}
    \setlength{\itemsep}{-0.3em}
    \item\label{instr} The man saw the woman \underline{with the telescope}.
    \item\label{NPmodifier} The man saw the woman \underline{with the dress}.
    \item\label{highAttach} The man saw the woman \underline{with his own eyes}.
\end{enumerate}

Even though sentences of the form \textit{NP V'ed NP with NP} are always \textit{structurally} ambiguous between an interpretation where the PP modifies the object NP -- the so-called {\em low attachment} reading exemplified by~\ref{NPmodifier} -- and a reading where the PP is the instrument of the VP -- the {\em high attachment} reading exemplified by~\ref{highAttach} --  our world knowledge can help us resolve such ambiguities by ruling out unlikely interpretations. In particular, although it is arguably roughly equally conceivable to do both interpretations in~\ref{instr} rendering both structurally and semantically ambiguity, in the absence of licensing context, the sentences in~\ref{NPmodifier} and~\ref{highAttach} are semantically unambiguous, because it is implausible for someone to use a dress as a seeing device, and implausible for someone to walk around with another person's eyes. Once supporting information is introduced, however, implausible sentences can become plausible~\cite{nieuwland2006peanuts}. For example, a context in which someone calls a telescope a dress or mistakes a telescope for a dress or has a dress-shaped telescope can make the instrument-PP interpretation in~\ref{NPmodifier} more likely, revealing the sentence's underlying structural ambiguity. In the absence of such licensing context, reading time studies show that humans have a general preference to attach an ambiguous PP to the VP rather than the NP (high attachment over low attachment), with higher reading times around the PP region, when the high attachment reading is {\em not} semantically licensed~\citep{spivey1995resolving}, such as in~\ref{NPmodifier}.

\begin{figure}
    \centering
    \includegraphics[width=0.95\columnwidth]{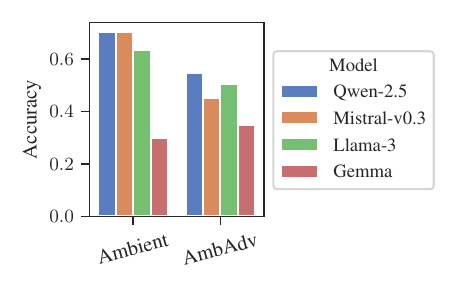}

    \caption{Results across models for an existing ambiguity dataset, Ambient, and our adversarial ambiguity dataset, AmbAdv.}
    \label{fig:fig1}
    \vspace{-7pt}
\end{figure}

Ambiguity in language poses a great challenge for Language Models (LMs) ~\citep[see Figure~\ref{fig:fig1}; ][]{liu-etal-2023-afraid}, as it can lead to hallucinations, biased completions, and misinterpretations. This can degrade performance on tasks such as fact-checking, sentiment analysis, and information extraction~\citep{10825265}. Different types of ambiguity affect models in distinct ways: syntactic ambiguity (e.g., \ref{instr}) can lead to parsing errors and flawed summarization or translation; lexical ambiguity (e.g., “the speaker”) can confuse question answering and retrieval systems, especially in low-context scenarios; and phonological ambiguity (e.g., “ice cream” vs. “I scream”) complicates speech recognition and dialogue modeling. Models that can detect and reason about ambiguity can help mitigate these issues by flagging unclear inputs, suggesting alternative interpretations, and improving user understanding—particularly in sensitive contexts like misleading news headlines or political claims\citep{liu-etal-2023-afraid}. Ambiguity-aware modeling is therefore essential for generating coherent, trustworthy outputs that align with human expectations~\citep{kamath-etal-2024-scope}.

To systematically evaluate how well LMs handle ambiguity, we introduce AmbAdv, a dataset with adversarial variations of ambiguous sentences, and find that most LLMs struggle to identify ambiguity, with some exhibiting a `yes bias' when prompted, highlighting the need to account for class distribution in evaluation. We examine how LLMs encode ambiguous sentences and find that linear probes on layer-wise representations can reliably distinguish ambiguity. Our analysis of model disambiguations and representations suggests that LLMs may partially rely on memorization to resolve ambiguity.

\section{Related Work}

Ambiguity enables efficient communication by relying on context and minimizing processing load~\citep{Piantadosi2011-cs}. Previous works have already highlighted ambiguity as a challenge in a variety of NLP tasks such as multimodal machine translation~\citep{li-etal-2022-visa}, visual question answering~\citep{stengel-eskin-etal-2023-chicken}, 
misleading claim detection~\citep{liu-etal-2023-afraid}
speech-to-text transcription
\citep{zhu2024resolving}, humour style classification~\citep{kenneth2024systematic}, sentiment analysis~\citep{buscemi2024chatgpt}, semantic parsing~\citep{stengel-eskin2024zero}, etc. There exist a few attempts to create ambiguity-inclusive datasets.~\citet{min-etal-2020-ambigqa} introduce AMBIGQA, identifying all possible answers to questions and rephrasing/disambiguating them.~\citet{liu-etal-2023-afraid} create AMBIENT, an NLI benchmark for ambiguity detection and disambiguation.~\citet{scopeambig} create a dataset of 1,000 scope-ambiguous sentences, showing that models may be sensitive to the meaning ambiguity. 

Building on these challenges, we construct an adversarial ambiguity dataset to evaluate whether models can detect ambiguity in inputs resembling real-world user queries. While most prior datasets focus on lexical or syntactic ambiguity, phonological ambiguity remains underexplored. Recent speech modeling studies show how transformer-based models like Whisper and Wav2Vec2 handle phonological variation, including homophones~\citep{mohebbi-etal-2023-homophone}, phonotactic patterns~\citep{deheerklootsHumanlikeLinguisticBiases2024}, and assimilation~\citep{pouw-etal-2024-perception}. These findings suggest that neural models can encode ambiguity-relevant distinctions in the speech domain, motivating our inclusion of phonological ambiguity and rhyme-based perturbations. We position AmbAdv relative to existing datasets in Table~\ref{tab:dataset-comparison} and provide further motivation in Appendix~\ref{sec:motivation}.

\section{Methodology}
\paragraph{Dataset creation}
We construct a dataset based on $8$ syntactically, $16$ lexically, and $16$ phonologically ambiguous sentences. These sentences were hand-picked based on existing textbooks, online linguistic studies, and ambiguity datasets~\citep{Taha_syntactic, liu-etal-2023-afraid, stengel-eskin2024zero}. 

\begin{table}[ht!]
\centering
\resizebox{0.9\columnwidth}{!}{%

\begin{tabular}{lccc}
\toprule
\textbf{Modification Type} & \textbf{Syntactic} & \textbf{Lexical} & \textbf{Phonological}\\ 
\midrule
Original   & 8    & 16  & 16  \\ 
Word Order    & 8    & 16   & -  \\ 
Synonym    & 540  & 64  & 64  \\ 
Random     & 404  & 64  & 64  \\ 
Rhyme      & 137  & 64  & 64  \\ 
\midrule
\textbf{Total} & 1097 & 224 & 208 \\ 
\bottomrule
\end{tabular}
}
\caption{Count of adversarial modifications across different linguistic ambiguity types.}
\label{tab:base-stats}
\end{table}

We modified the originally ambiguous base sentences to create sentence variants each using four different manipulation types:
(i) \textit{Word order.} We swap the subject and the object of the sentence or, when not possible, create passive/active voice alternations. Critically, this manipulation does not lead to a semantically implausible interpretation of either of the PP attachment possibilities.
(ii) \textit{Synonymous word.} We exchange a key word in the sentence for a synonym. Because we use synonyms, these sentence variants have the same ambiguity structure as the original example.
(iii) \textit{Random word.} We exchange a keyword in the sentence for a random word of the same syntactic category. These sentence variants rule out one interpretation of the sentence (i.e., here: the PP can no longer be interpreted as introducing the instrument of the seeing action). (iv) \textit{Rhyme word.} We exchange a key word in the sentence for a word that rhymes in sound with the original word. 

Table~\ref{tab:data-table} shows an example of these data manipulations, and Table~\ref{tab:base-stats} presents the dataset statistics. The total number of sentences in AmbAdv is $1529$, with $671$ labeled as ambiguous. Further details can be found in Appendix~\ref{sec:base-sentences}. Each manipulation type targets a different type of ambiguity and model robustness. Word substitutions have long been used in adversarial NLP~\citep{zhou-etal-2021-defense, 10.1145/3593042, bespalov-etal-2023-towards, sabir2023interpretability, 10646716} to explore model vulnerabilities and assess generalization under distribution shifts. Word order can affect model performance \citep{abdou-etal-2022-word}. Synonym substitutions simulate natural lexical variation while preserving ambiguity, testing semantic resilience~\citep{hsieh-etal-2019-robustness}. Random substitutions disrupt semantic coherence to test model reliance on lexical cues. Rhyme-based substitutions, introduced here as a novel perturbation, explore phonological similarity, an underexplored adversarial strategy in NLP~\citep{suvarna-etal-2024-phonologybench}. This is particularly relevant given the growing interest in phonological effects in LLM outputs, especially in multimodal contexts~\citep{fathullah-etal-2024-audiochatllama}. 

\begin{table}[ht!]
\resizebox{\columnwidth}{!}{%
\begin{tabular}{llcc}
\toprule
& \multirow{2}{*}{} & \multicolumn{2}{c}{\textbf{Ambiguous?}} \\
\cmidrule(lr){3-4}
\textbf{Manipulation} & \textbf{Example sentence}   & \textbf{Struct.} &  \textbf{Semant.}              \\ \midrule
Original             & The man saw the woman with the telescope. & \cmark & \cmark  \\
Word Order         & The \textit{woman} saw the \textit{man} with the telescope.  & \cmark & \cmark \\
Synonym Word        & The man saw the woman with the \textit{binoculars}. & \cmark & \cmark \\
Random Word         & The man saw the woman with the \textit{book}.     & \cmark & \xmark  \\
Rhyme Word          & The man saw the woman with the \textit{gyroscope}. & \cmark & \xmark \\

\bottomrule
\end{tabular}%
}
\caption{Overview of manipulation types for the syntactically ambiguous sentences.}
\label{tab:data-table}
\end{table}

\begin{figure*}[h]
    \centering
    \includegraphics[width=0.85\textwidth]{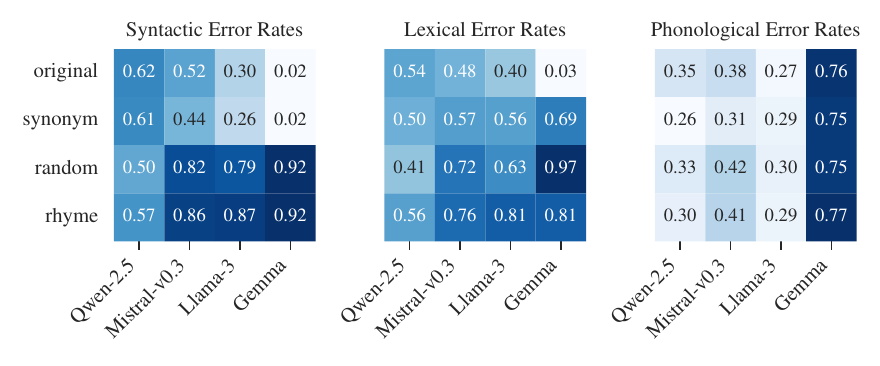}
    \vspace{-7pt}
    \caption{Results across models after zero-shot prompting, showing the error rate of the different synonym, random, and rhyme word modifications. Values closer to 1 indicate a higher error rate.}
    \label{fig:heatmap}
    \vspace{-7pt}
\end{figure*}
\begin{table*}[ht]
    \centering
    \resizebox{0.63\textwidth}{!}{%
    \begin{tabular}{lcccc}
    \toprule
    {} & Syntactic & Lexical & Phonetic & Ambient \\
    \midrule
    Qwen-2.5 & 0.436 \footnotesize{$\pm$ 0.028} & 0.491 \footnotesize{$\pm$ 0.128} & 0.781 \footnotesize{$\pm$ 0.032} & 0.693 \footnotesize{$\pm$ 0.025} \\
    Mistral-v0.3 & 0.360 \footnotesize{$\pm$ 0.064} & 0.407 \footnotesize{$\pm$ 0.088} & 0.647 \footnotesize{$\pm$ 0.197} & 0.703 \footnotesize{$\pm$ 0.055} \\
    Llama-3 & 0.462 \footnotesize{$\pm$ 0.026} & 0.360 \footnotesize{$\pm$ 0.058} & 0.697 \footnotesize{$\pm$ 0.105} & 0.636 \footnotesize{$\pm$ 0.068} \\
    Gemma & 0.525 \footnotesize{$\pm$ 0.015} & 0.268 \footnotesize{$\pm$ 0.006} & 0.247 \footnotesize{$\pm$ 0.092} & 0.299 \footnotesize{$\pm$ 0.080} \\
    \bottomrule
    \end{tabular}
    }
    \caption{Average accuracy results across 8 different prompt templates.}
    \label{tab:tab1}
\end{table*}

\paragraph{Dataset validation}
Lexical and phonological ambiguities were straightforward to annotate and validate. Lexical ambiguity was verified via dictionary lookup, while phonological ambiguity was confirmed using IPA transcriptions from authoritative sources to identify homophones or rhyming patterns. On the contrary, the syntactic sentences are all syntactically ambiguous, allowing multiple interpretations of their structure. However, considering the meanings of the words can often disambiguate them. For example, ``\emph{The man saw the woman with the dress}'' (a random word substitution) is structurally ambiguous, but our world knowledge precludes interpreting a dress as a visual instrument.\footnote{Except, of course, if the dress has a mirror. Such exceptions illustrate how pragmatics may override semantics~\cite{Morris1946-MORSLA-2}, but LLMs must be sensitive to distinctions between syntactic ambiguities that support multiple conventional readings and those typically resolved by context.} Based on this extra-linguistic reasoning, we classified sentences with random and rhyme word substitutions as \emph{unambiguous}. 

To validate our annotations, three annotators independently labeled a random 50\% sample of the dataset as ambiguous or unambiguous, incorporating real-world knowledge in their judgments. Inter-annotator agreement, measured by Cohen’s Kappa, was $91\%$. Agreement for synonym-based sentences was \circa$84\%$, and for random/rhyme substitutions, around \circa$98\%$.

\paragraph{Models}
We are interested in evaluating whether LLMs can reliably modulate their judgments of a sentence's ambiguity status in the face of minimal adversarial attacks that either change or do not change the sentence's ground truth ambiguity label. To this end, we use four open-access instruction-tuned LLMs in the $7$ Billion parameter regime, selected based on their performance on the LMSys chatbot arena leaderboard:\footnote{\huggingface{https://huggingface.co/spaces/lmsys/chatbot-arena-leaderboard}} \texttt{Qwen-2.5-7b}, \texttt{Mistral-7b-v0.3}, \texttt{Llama-3-7b}, and \texttt{Gemma-7b}. We set the experiments in a binary classification set-up because it provides a straightforward and interpretable approach, which can also be seen in a real-world scenario in which a user asks a model if a sentence (e.g., a news headline) is ambiguous. 

\paragraph{Prompting strategy}
We frame our experiments as ambiguity identification tasks. We use 8 different prompts that elicit responses in a structured Jinja2 format across the different model chat templates.\footnote{\fngithub{https://github.com/jndiogo/LLM-chat-templates}} The prompts are: 2 default prompts and 6 prompts with various binary words in different orders to investigate a potential ``yes-bias''~\citep{doi:10.1073/pnas.2309583120} and model consistency,\footnote{We define consistency here as the ability \textit{to make reliable decisions in semantically similar contexts} demonstrating a systematic capacity to generalize across language variations"~\citep{elazar-etal-2021-measuring}.}. For a set of 120 sentences, we also use 2 disambiguation prompts. We provide the templates in Appendix~\ref{app:prompt_templates}. To evaluate the model responses, we calculate the accuracy/error rate, i.e., the proportion of matches/mismatches between the annotated and predicted ambiguity status.

\section{Experiments and Results}

\paragraph{Ambiguity Identification}

We visualize the average results over five runs across 8 prompts in Figure~\ref{fig:fig1} and Table~\ref{tab:tab1}. We observe several trends: 
\textbf{(1)} Most models perform worse on AmbAdv, even though the sentences are simpler compared to Ambient, indicating some model sensitivity in the adversarial sentence modifications. \textbf{(2)} The models demonstrate consistent performance across 8 different prompts, with an average accuracy difference of less than 0.1 points. \textbf{(3)} Qwen and Mistral perform worst in the syntactically ambiguous sentences, while Llama performs worst in the lexical ambiguous sentences. \textbf{(4)} Gemma performs better on the syntactic set, but upon checking the distribution of responses, we found that it gives almost all affirmative answers. The syntactic set is more balanced with equal positive and negative examples, while the other datasets contain more negative examples (see Figure~\ref{fig:fig-yes-no}). \textbf{(5)} This suggests a “yes bias” in most of the model outputs. 

\begin{figure*}[h!]
    \centering
    \includegraphics[width=\textwidth]{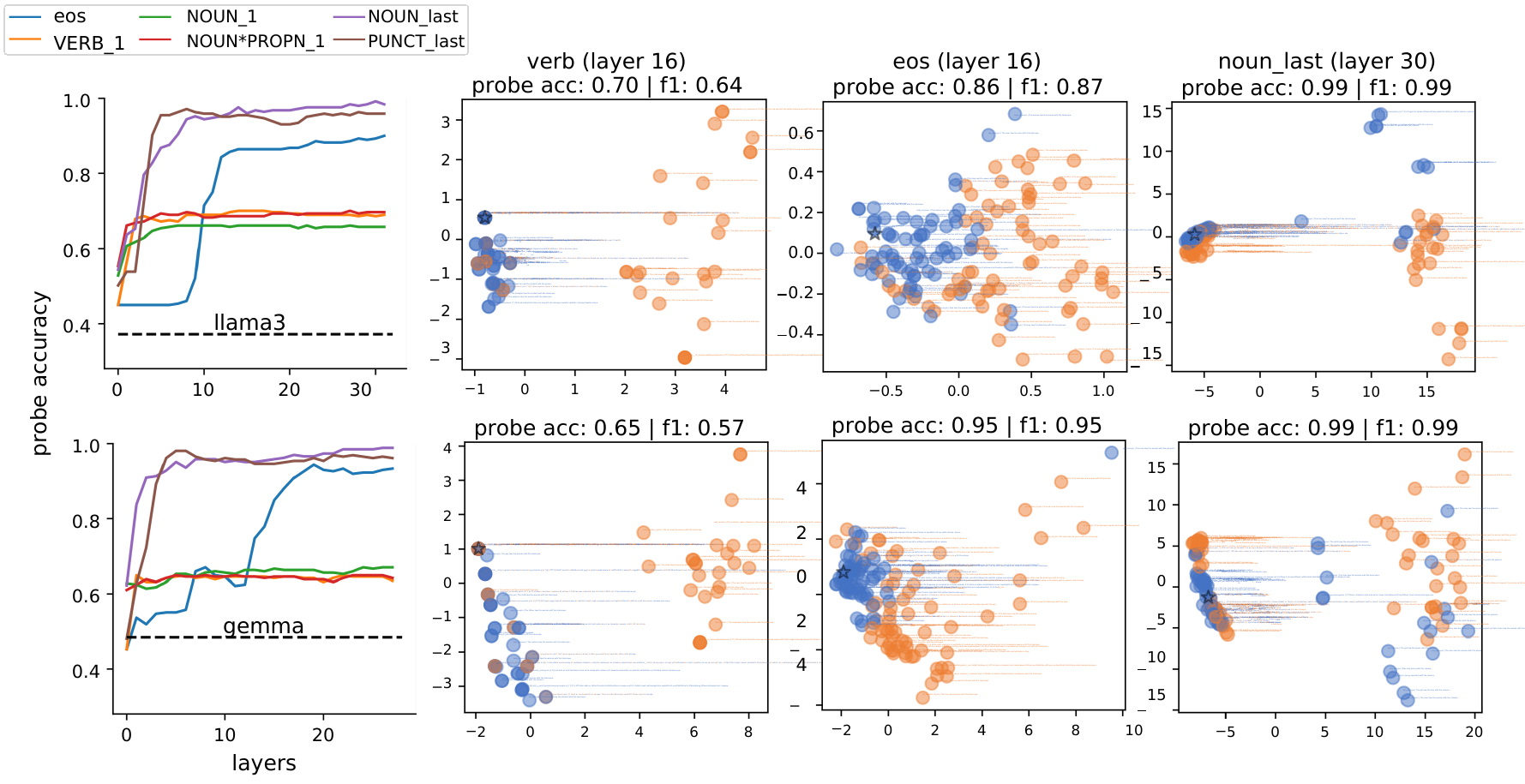}
    \caption{Representational analysis of hidden state representation across layers. Linear Probe accuracies for Llama-3 and Gemma for different functional roles of tokens, e.g., first verb or last noun (left). PCA projections extracted for different token roles, blue/orange indicate ambiguous/unambiguous sentences (right).}
    \label{fig:rep}
    \vspace{-7pt}
\end{figure*}

\paragraph{Adversarial Sentence modifications.}
We provide an aggregation of each model's performance across types of sentence modifications in Figure~\ref{fig:heatmap}. We observe the following: \textbf{(1)} Ambiguity identification seems to be a challenging task for most models. \textbf{(2)} Synonym words are not always more challenging than the original sentences. \textbf{(3)} Both random and rhyming words pose the greatest challenge for most models, suggesting a lack of resilience to adversarial substitutions. \textbf{(4)} Qwen is the only model that performs better on random substitutions and struggles with the original sentences. This behavior may suggest an attempt to avoid memorization of the original sentences, which were likely part of its training data. Alternatively, it could indicate that the model maintains a more balanced approach in generating yes/no responses.

\paragraph{Disambiguation analysis}
For the syntactic ambiguity set, we prompted models to also provide disambiguations on a set of 120 sentences. Manual error analysis revealed the following: \textbf{(1)} Only Qwen and Llama provided accurate disambiguations, but only for the original and synonym cases. Mistral and Gemma add hallucinations in their responses (e.g., `The man saw a woman near the location of the dress', or `The man saw the woman with the camera, and she was taking pictures').  \textbf{(2)} Despite identifying ambiguity, many models lack world knowledge—e.g., they change the action verb with synonym instruments for `telescope' (e.g., `using or holding a camera/glasses/polaroid' but not `seeing with a camera/glasses/polaroid'). Additionally, inanimate objects are sometimes assigned actions (e.g., `A mug witnessed a woman using a telescope', `The telescope saw the man who was with the woman').  \textbf{(3)} In some cases, the model repeated the disambiguation of the original sentence in synonym substitutions, indicating a memorization effect.
We provide a sample of the model responses in the Appendix Tables~\ref{tab:dis} and~\ref{tab:dis-2}.

\paragraph{Representational Analysis}

After assessing the behavioural model performance and error patterns, we now investigate to what extent residual stream representations can decode ambiguity in a subset of 142 samples of the AmbAdv dataset that contains variations of a sentence~\ref{instr}. We extract layer-wise representations of specific token roles, i.e., first verb, first noun/proper noun, last noun, last punctuation, and the final end of prompt sentence representation (eos). We compute probe accuracies \cite{radford2021learning, campbell2023localizing} for each role via a 5-fold linear probe evaluation using a logistic regression model (80/20 train/test splits). 
We present a summary of our results in Figure~\ref{fig:rep}, with additional details provided in Appendix~\ref{app:rep_analysis}. Our findings show that averaged probe accuracies are consistently higher compared to the performance of the model's generated responses through prompting. We find probe accuracies of $0.9$ and higher for token representations related to the last punctuation and last noun token, starting from layer 5 on. The first noun/proper noun and verb reach moderately high accuracy scores of $0.6-0.7$. The final prompt token, used by the language modeling head to generate the answer token, achieves probe accuracies of $0.85-0.90$ from layer 12 on, which appear delayed compared to the other token types considered. An analysis of PCA projections of representations highlights that ambiguous/unambiguous (blue/orange in Figure~\ref{fig:rep} samples appear clustered, yet we do not find a clear direction that would uniquely code for ambiguity. Interestingly, we observe groups of both ambiguous and unambiguous sentences clustering around the original sentence (indicated by a blue star). This offers representational insight into repeated misclassifications, which may stem from memorized patterns, as many AmbAdv source sentences are publicly available online.

\section{Conclusion}

We introduce the first adversarial dataset for linguistic ambiguity to evaluate whether LLMs can assess a sentence's ambiguity status under minimal adversarial attacks that may or may not alter its ground truth label. Our results show that LLMs struggle to accurately interpret ambiguous sentences. While ambiguity-related information seems present in the models' representations, they fail to leverage it effectively in their outputs. Our representational analysis shows that linear probes on layer-wise representations can reliably distinguish ambiguity, suggesting that LLMs may partially rely on memorization rather than ambiguity understanding.

\section{Limitations}

We use prompting as our evaluation paradigm to explore the model and prompting limitations, as this method is likely to be used by users of language models in real-world settings. For example, a user might ask if a given sentence (e.g., a news headline) is ambiguous. This is why we designed the dataset with different types of linguistic ambiguity sentences, aiming to systematically investigate how well models can identify ambiguity even in publicly available and adversarial examples. Before the LLM era, linguistic knowledge encoded in neural language models and LLMs has been evaluated using either log likelihood comparisons of minimal pair sentences \cite{linzen-2016-assessing, futrell-etal-2019-neural, warstadt-2020-blimp, hu-etal-2020-systematic, hu2024language} or through probing of the model's representations of a stimulus \cite{hewitt-manning-2019-structural, eisape-etal-2022-probing, muller-eberstein-etal-2022-probing}. Similarly, semantic plausibility has been evaluated using log-likelihood measures and representation probing \cite{kauf2023event, michaelov2023can, misra2024experimental}. 

More recently, \textit{prompting} emerged as a way to directly prompt LLMs for linguistic knowledge using natural language (e.g., \citealp{brown2020language,blevins-etal-2023-prompting}). Nevertheless, a direct comparison of log-likelihood and prompting measures shows that prompting may systematically underestimate the model's true linguistic capabilities because it requires the models not only to solve the task, but also to correctly interpret the prompt and to translate their answer into the desired output format \cite{hu-levy-2023-prompting, hu2024language}. This is the reason that motivated us to include a representational analysis, in addition to a prompting-type analysis. 

Moreover, adversarial datasets, having been specifically designed to fool a model, may be subject to the biases of the dataset creators~\citep{li-michael-2022-overconfidence}. However, the purpose of our dataset is not to train and build models that are more robust to spurious correlations but rather to interpret and evaluate certain model behaviours. The size of our dataset may also be a limitation, as it can only be used for cases that evaluate the sensitivity of LMs toward ambiguity and adversarial examples. Lastly, a major limitation is that our dataset only includes English sentences, which limits its applicability to other languages. We provide further motivation for our study and dataset choices in Appendix~\ref{sec:motivation}.

\section{Ethics}

We do not foresee any ethical concerns. On the contrary, the scope of our dataset is purely for scientific research of language models and can potentially help identify ambiguity in political claims, news headlines, and other domains. The research was conducted in accordance with ethical principles, and no sensitive or personal data was used or collected during the study.  We fairly compensated each annotator involved in this study at a rate of \$$18$ per hour on the crowdsourcing platform Prolific.\footnote{\url{https://www.prolific.com/}} The only requirement for annotators' demographic and geographic characteristics was being a native English speaker. The instructions given to the annotators were very similar to the default prompts we used in our work (see Appendix \ref{app:prompt_templates}. The dataset contains linguistic sentences that can be found in grammar books and do not raise any privacy or ethical concerns. In terms of resources we estimate less than 12 hours GPU (A100~40GB) usage. Both our dataset and code are publicly available under CC BY 4.0 \footnote{\url{https://creativecommons.org/licenses/by/4.0/}} and MIT licences respectively at \github{https://github.com/coastalcph/lm_ambiguity}.

\appendix

\section{Motivation}
\label{sec:motivation}

The motivation behind creating AmbAdv, the first adversarial ambiguity dataset, lies in the fact that ambiguity remains a persistent challenge for language models. Prior work has explored ambiguity in narrow contexts, such as natural language inference (NLI)~\citep{liu-etal-2023-afraid}, semantic parsing~\citep{stengel-eskin2024zero}, and log probabilities~\citep{kamath-etal-2024-scope}. These studies primarily focus on syntactic and scope ambiguity, assessing how LLMs handle ambiguous inputs in zero-shot and few-shot settings. However, previous work has not directly assessed LLMs' ability to detect ambiguity through user-facing interactions. Since users engage with LLMs via direct prompts, it is crucial to understand how models respond to ambiguous inputs \texttt{in real-world usage}.

To systematically evaluate ambiguity sensitivity, we focus on three key ambiguity types: syntactic, lexical, and phonological. This selection is grounded in cognitive and linguistic theories~\citep{amb}, which classify ambiguity into lexical, phonological, morphological, and syntactic categories. We put a great focus on syntactic ambiguity, particularly PP attachment ambiguities, as it poses challenges for parsing and can lead to garden-path effects. Moreover, identifying between the two possible structures/readings of a syntactically ambiguous sentence often requires background world knowledge~\citep{synt_motiv}, which is a challenging concept for an artificial intelligence model~\citep{ivanova2024elements}. We included lexical ambiguity, as it is the most frequent, as word meanings are highly flexible and influenced by both linguistic and extralinguistic factors. Lastly, phonological ambiguity, though rarely explored in computational settings, is crucial in spoken language, humour, and wordplay—yet, to our knowledge, no dataset has included it. Morphological ambiguity was not included, as it was infeasible to systematically construct adversarial variations.

By focusing on these ambiguity types, we aim to broaden the scope of ambiguity evaluation and examine whether LLMs can recognize and handle the diverse sources of linguistic uncertainty that shape human communication.

\begin{table*}[!ht]
\centering
\resizebox{\textwidth}{!}{%
\begin{tabular}{@{}lcccc@{}}
\toprule
\textbf{Dataset} & \textbf{Size} & \textbf{Ambiguity Types} & \textbf{Use Case} & \textbf{Adversarial} \\
\midrule
AmbigQA~\citep{min-etal-2020-ambigqa} & 14,042 & Referential & Open-domain QA & \xmark \\
Ambient~\citep{liu-etal-2023-afraid} & 1,645 & Lexical, Syntactic & NLI, Disambiguation & \xmark \\
AMP~\citep{stengel-eskin2024zero} & Synthetic & Lexical, Syntactic & Semantic Parsing & \xmark \\
Scope~\citep{kamath-etal-2024-scope} & 1,000 & Scope & Human Preference Analysis & \xmark \\
\textbf{AmbAdv (Ours)} & 1,529 & Syntactic, Lexical, Phonological & Ambiguity Identification, Disambiguation & \checkmark \\
\bottomrule
\end{tabular}
}
\caption{Comparison of AmbAdv with related ambiguity datasets in terms of size, ambiguity types, intended use case, and adversarial construction.}
\label{tab:dataset-comparison}
\end{table*}

\subsection{Comparison with Prior Datasets}

While several recent datasets have explored ambiguity in language models, AmbAdv is unique in its adversarial construction and coverage of ambiguity types. Unlike prior resources such as AmbigQA, Ambient, AMP, and Scope, which primarily focus on naturally occurring ambiguities within specific tasks like open-domain question answering or natural language inference, AmbAdv systematically introduces controlled perturbations—such as synonym replacements, random substitutions, and rhyming alterations—to create ambiguous and unambiguous sentence pairs. This adversarial approach is designed to rigorously evaluate model robustness in handling ambiguity. Additionally, AmbAdv is the first to incorporate phonological ambiguity (i.e., homonyms) into its evaluation framework, addressing a previously underexplored area in ambiguity research. By centering on the task of ambiguity identification, AmbAdv challenges models to detect and interpret ambiguous inputs without external context, thereby complementing existing resources and providing a valuable benchmark for advancing research in ambiguity detection and resolution within NLP systems. Table~\ref{tab:dataset-comparison} provides a summary that can help position our dataset relative to prior work.

\section{Dataset Details}
\label{sec:base-sentences}

The dataset was curated by the authors of the paper after collecting linguistic ambiguity sentences from various online linguistic textbooks and publicly available datasets~\citep{Taha_syntactic, liu-etal-2023-afraid, stengel-eskin2024zero}. For syntactic ambiguity, we generated a total of 1,097 sentences, with sentence variants for each manipulation type and part of speech. We decided to provide multiple variations for the different parts of the sentence, as the set of sentences has a similar structure, and we wanted to reflect how ambiguity arises from a network of heterogeneous participants—agents, subjects, and objects—each with varying roles~\citep{amb}. For lexical and phonological sentences, we created 4 examples per manipulation type, resulting in a total of 224 sentences for lexical ambiguity and 208 for phonological ambiguity (we did not have a different word order in phonological ambiguity because the homonyms changed). 

Many researchers have explored word substitution as a form of adversarial attack in NLP~\citep{zhou-etal-2021-defense, bespalov-etal-2023-towards, sabir2023interpretability, 10646716}. Building on this foundation, we selected synonym replacement, word order changes, and random perturbations as core transformation types, as these are commonly used in adversarial defense benchmarks and robustness evaluations~\citep{10.1145/3593042}. To extend beyond traditional lexical and syntactic manipulations, we introduced a novel rhyme-based perturbation, motivated by the underexplored area of phonological ambiguity. This choice is further supported by the growing interest in phonological effects in LLM outputs~\citep{fathullah-etal-2024-audiochatllama, suvarna-etal-2024-phonologybench}, particularly as models increasingly integrate multimodal capabilities. By incorporating this diverse set of perturbations, our study aims to investigate a broader spectrum of ambiguity types and assess model robustness across both well-studied and emerging linguistic phenomena.

Rhyme substitutions explore phonological similarity, an underexplored adversarial strategy in NLP~\citep{suvarna-etal-2024-phonologybench}. These are particularly relevant given the increasing attention to phonological effects in LLM outputs, especially in spoken or dialogue contexts~\citep{fathullah-etal-2024-audiochatllama}.

Table~\ref{tab:base-sentences} presents the complete set of base sentences used for syntactic ambiguity. Table~\ref{tab:phonological-ambiguity} lists all base sentences for phonological ambiguity, and Table~\ref{tab:lexical-ambiguity} provides the full set for lexical ambiguity.

\begin{table}[ht!]
\resizebox{\columnwidth}{!}{%
\begin{tabular}{ll}
\toprule
\textbf{No.} & \textbf{Sentence} \\ \midrule
1 & The man saw the woman with the telescope. \\
2 & She fed her cat food. \\
3 & Harry loves his pet turtle more than his wife. \\
4 & The captain ordered the old men and women of the ship. \\
5 & I saw a dog in my pyjamas. \\
6 & An enraged cow injured a farmer with an ax. \\
7 & The hospital is being sued by six foot doctors. \\
8 & Helen got lunch ready for her daughter wearing a summer dress. \\
\bottomrule
\end{tabular}
}
\caption{Set of syntactically ambiguous base sentences.}
\label{tab:base-sentences}
\end{table}

\begin{table}[ht!]
\resizebox{\columnwidth}{!}{%
\begin{tabular}{ll}
\toprule
\textbf{No.} & \textbf{Sentence Pair} \\ \midrule
1 & It’s not easy to wreck a nice beach. / It’s not easy to recognize speech. \\
2 & I saw a sea monster. / I saw a seam on stir. \\
3 & She sells seashells by the seashore. / She sells sea shells buy the sea sure. \\
4 & Whale meat again. / We'll meet again. \\
5 & I saw a bear in the forest. / Eye sore a bare inn the for rest. \\
6 & He took a nice cold shower after his date. / He took an ice cold shower after his date. \\
7 & The stuffy nose is annoying. / The stuff he knows is annoying. \\
8 & He couldn't wait to leave the mall. / He couldn't wait to leave them all. \\
9 & The good can decay many ways. / The good candy came anyways. \\
10 & She can't bear to lose the race. / She can't bear to lose their ace. \\
11 & He's a man of many talents. / He's a man of mini talents. \\
12 & I hurt myself with the four candles. / I hurt myself with the fork handles. \\
13 & I ordered pepperoni pizza. / I ordered pepper only pizza. \\
14 & They wanted to explore the ancient ruins. / They wanted to explore the agent's ruins. \\
15 & I love you. / Aisle of view. \\
16 & Caesar salad. / Seize her salad. \\
\bottomrule
\end{tabular}%
}
\caption{Set of phonologically ambiguous sentence pairs.}
\label{tab:phonological-ambiguity}
\end{table}

\begin{table}[ht!]
\resizebox{0.7\columnwidth}{!}{%
\begin{tabular}{ll}
\toprule
\textbf{No.} & \textbf{Sentence} \\ \midrule
1 & Your boss is a funny man. \\
2 & The speaker is at the front of the room. \\
3 & He's not very well off. \\
4 & John and Anna are married. \\
5 & It is my belief that the earth is round. \\
6 & She is looking for a match. \\
7 & Give me a ring. \\
8 & Show me the light feathers. \\
9 & We saw her duck. \\
10 & I'm going to take a break from studying. \\
11 & Alice and Jon disagreed. \\
12 & Give me the bat! \\
13 & What you said is insane. \\
14 & Yesterday I went to the bank. \\
15 & I can't find the glasses. \\
16 & He didn't see the picture of the disaster. \\
\bottomrule
\end{tabular}%
}
\caption{Set of lexically ambiguous sentences.}
\label{tab:lexical-ambiguity}
\end{table}

To ensure consistency in our annotations, we applied distinct validation strategies for each ambiguity type. Lexical ambiguity was identified through dictionary lookup, confirming that a word had multiple meanings depending on context. Phonological ambiguity was validated using IPA transcriptions from authoritative linguistic resources to verify homophony or rhyming patterns. In contrast, syntactic ambiguity was treated differently: all syntactic examples were constructed to allow multiple structural interpretations. However, we observed that many such cases could be pragmatically disambiguated based on world knowledge. For instance, in the sentence ``\emph{The man saw the woman with the dress},'' although the syntax permits multiple parses, common sense rules out interpreting the dress as a visual instrument. Based on this reasoning, we labeled sentences with random or rhyme-based substitutions as \emph{unambiguous}, since their ambiguity was not plausible under typical interpretive contexts.

All annotators were recruited via the Prolific crowdsourcing platform\footnote{\url{https://www.prolific.com/}} and compensated fairly at a rate of \$$18$ per hour. The only eligibility criterion was being a native English speaker, with no restrictions on geographic location. Annotators received task instructions closely aligned with the default prompts used in our experiments (see Appendix~\ref{app:prompt_templates} for details), ensuring consistency between model and human evaluations.

\begin{figure*}
    \centering
    \includegraphics[width=0.9\textwidth]{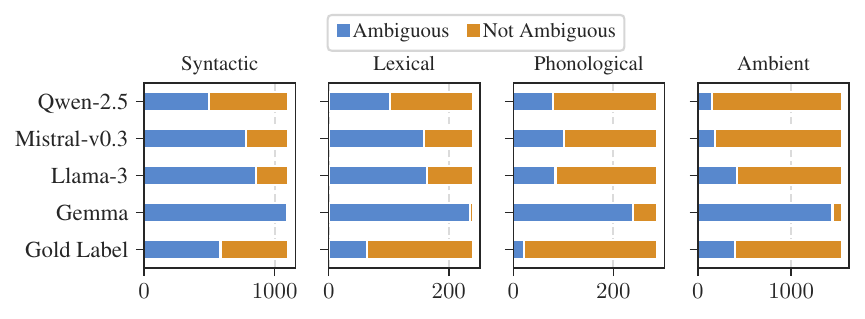}
    \caption{Comparison of the ambiguous and non-ambiguous count of sentences predicted by language models against the 'gold standard label'.}
    \label{fig:fig-yes-no}
\end{figure*}

We also present in Figure~\ref{fig:fig-yes-no} a distribution of the ambiguous or non-ambiguous sentences comparing the \textit{Gold Label} as annotated in the datasets, and the model predictions.

\section{Experiments}
\subsection{Prompt Templates}
\label{app:prompt_templates}

We used a total of 8 different templates for ambiguity identification: 2 default templates -one asking directly if the sentence is ambiguous and one asking indirectly if the sentence has two interpretations-, and 6 templates with a binary word alteration between true, false, yes, no, right, and wrong. We also used 2 templates for ambiguity disambiguation, which are modified versions of the default templates. We provide indicative examples of the Llama with the llama chat template in Figures~\ref{fig:prompts}-\ref{fig:prompts-bin-pho}. All templates and model outputs will be available upon releasing our codebase.

\begin{figure*}[tb!]
\begin{tcolorbox}[
    colback=white,
    colframe=blue
]
\begin{verbatim}
    
<|start_header_id|>system<|end_header_id|>

Considering real-world knowledge, could this sentence be interpreted in more than 
one way?

{{ sentence }}

Format your response according to the following JSON schema:
{{ schema }}
<|eot_id|>
\end{verbatim}
\end{tcolorbox}
\caption{Default1 prompt used for identifying syntactic and lexical ambiguities.}
\label{fig:prompts}
\end{figure*}

\begin{figure*}[tb!]
\begin{tcolorbox}[
    colback=white,
    colframe=blue
]
\begin{verbatim}
    
<|start_header_id|>system<|end_header_id|>

Based on general world knowledge, is the following sentence ambiguous?

{{ sentence }}

Format your response according to the following JSON schema:
{{ schema }}
<|eot_id|>
\end{verbatim}
\end{tcolorbox}
\caption{Default2 prompt used for identifying syntactic and lexical ambiguities.}
\label{fig:prompts2}
\end{figure*}

\begin{figure*}[tb!]
\begin{tcolorbox}[
    colback=white,
    colframe=green
]
\begin{verbatim}
    
<|start_header_id|>system<|end_header_id|>

Based on general world knowledge, is the following sentence ambiguous? 
Reply with True or False.

{{ sentence }}

Format your response according to the following JSON schema:
{{ schema }}
<|eot_id|>
\end{verbatim}
\end{tcolorbox}
\caption{Example of the True binary type alteration prompt for identifying syntactic and lexical ambiguities.}
\label{fig:prompts2-bin}
\end{figure*}

\begin{figure*}[tb!]
\begin{tcolorbox}[
    colback=white,
    colframe=green
]
\begin{verbatim}
    
<|start_header_id|>system<|end_header_id|>

Based on general world knowledge, is the following sentence ambiguous? 
Reply with False or True.

{{ sentence }}

Format your response according to the following JSON schema:
{{ schema }}
<|eot_id|>
\end{verbatim}
\end{tcolorbox}
\caption{Example of the False binary type alteration prompt for identifying syntactic and lexical ambiguities.}
\label{fig:prompts-bin}
\end{figure*}

\begin{figure*}[tb!]
\begin{tcolorbox}[
    colback=white,
    colframe=red
]
\begin{verbatim}
    
<|begin_of_text|><|start_header_id|>user<|end_header_id|>

Based on general world knowledge, is the following sentence ambiguous? If yes, 
provide the two disambiguations of the sentence separated by a comma.

{{ sentence }}

Format your response according to the following JSON schema:
{{ schema }}
<|eot_id|><|start_header_id|>assistant<|end_header_id|>
\end{verbatim}
\end{tcolorbox}
\caption{Example of the prompt used for the disambiguation task.}
\label{fig:prompts3}
\end{figure*}

\begin{figure*}[tb!]
\begin{tcolorbox}[
    colback=white,
    colframe=blue
]
\begin{verbatim}
    
<|start_header_id|>system<|end_header_id|>

Could these two sentences sound alike but have different meanings?

Sentence 1: {{ sentence1 }}
Sentence 2: {{ sentence2 }}

Format your response according to the following JSON schema:
{{ schema }}
<|eot_id|>
\end{verbatim}
\end{tcolorbox}
\caption{Default1 prompt used for identifying phonological ambiguity.}
\label{fig:prompts-pho}
\end{figure*}

\begin{figure*}[tb!]
\begin{tcolorbox}[
    colback=white,
    colframe=blue
]
\begin{verbatim}
    
<|start_header_id|>system<|end_header_id|>

Are these two sentences phonologically ambiguous?

Sentence 1: {{ sentence1 }}
Sentence 2: {{ sentence2 }}

Format your response according to the following JSON schema:
{{ schema }}
<|eot_id|>
\end{verbatim}
\end{tcolorbox}
\caption{Default2 prompt used for identifying phonological ambiguity.}
\label{fig:prompts2-pho}
\end{figure*}

\begin{figure*}[tb!]
\begin{tcolorbox}[
    colback=white,
    colframe=green
]
\begin{verbatim}
    
<|start_header_id|>system<|end_header_id|>

Are these two sentences phonologically ambiguous? Reply with True or False.

Sentence 1: {{ sentence1 }}
Sentence 2: {{ sentence2 }}

Format your response according to the following JSON schema:
{{ schema }}
<|eot_id|>
\end{verbatim}
\end{tcolorbox}
\caption{Example of the True binary type alteration prompt for identifying phonological ambiguity.}
\label{fig:prompts2-bin-pho}
\end{figure*}

\begin{figure*}[tb!]
\begin{tcolorbox}[
    colback=white,
    colframe=green
]
\begin{verbatim}
    
<|start_header_id|>system<|end_header_id|>

Are these two sentences phonologically ambiguous? Reply with False or True.

Sentence 1: {{ sentence1 }}
Sentence 2: {{ sentence2 }}

Format your response according to the following JSON schema:
{{ schema }}
<|eot_id|>
\end{verbatim}
\end{tcolorbox}
\caption{Example of the False binary type alteration prompt for identifying phonological ambiguity.}
\label{fig:prompts-bin-pho}
\end{figure*}

\subsection{Results across Prompts}
\label{subsec:results-prompts}
We present the results across the different prompt templates in Table~\ref{tab:ambiguity_results}. Overall we observe that models perform better on different prompts, and there does not seem to be an optimal prompt. The variations may be marginal in some cases, suggesting that most models we examined are not prompt-sensitive. The only model that seems to have a preference for a prompt is Gemma, which seems to prefer the default2 prompt asking directly if a sentence is ambiguous.

\begin{table*}
\centering
\subcaptionbox{Syntactic Ambiguity}[.9\linewidth]{
\begin{tabular}{lrrrrrrrrr}
\toprule
Model & Default1 & Default2 & False & No & Right & True & Wrong & Yes & AVG \\
\midrule
Qwen-2.5 & 0.409 & 0.409 & 0.473 & 0.479 & 0.433 & 0.444 & 0.426 & 0.414 & 0.436 \\
Mistral-v0.3 & 0.503 & 0.378 & 0.356 & 0.306 & 0.322 & 0.304 & 0.368 & 0.340 & 0.360 \\
Llama-3 & 0.511 & 0.486 & 0.440 & 0.473 & 0.447 & 0.450 & 0.439 & 0.447 & 0.462 \\
Gemma & 0.530 & 0.489 & 0.531 & 0.531 & 0.531 & 0.531 & 0.531 & 0.531 & 0.525 \\
\bottomrule
\end{tabular}
}

\vspace{1em}

\subcaptionbox{Lexical Ambiguity}[.9\linewidth]{
\begin{tabular}{lrrrrrrrrr}
\toprule
Model & Default1 & Default2 & False & No & Right & True & Wrong & Yes & AVG \\
\midrule
Qwen-2.5 & 0.354 & 0.354 & 0.608 & 0.588 & 0.550 & 0.617 & 0.542 & 0.312 & 0.491 \\
Mistral-v0.3 & 0.263 & 0.287 & 0.388 & 0.508 & 0.458 & 0.458 & 0.438 & 0.454 & 0.407 \\
Llama-3 & 0.271 & 0.438 & 0.400 & 0.375 & 0.375 & 0.404 & 0.333 & 0.287 & 0.360 \\
Gemma & 0.267 & 0.283 & 0.267 & 0.267 & 0.267 & 0.267 & 0.267 & 0.263 & 0.268 \\
\bottomrule
\end{tabular}
}

\vspace{1em}

\subcaptionbox{Phonetic Ambiguity}[.9\linewidth]{
\begin{tabular}{lrrrrrrrrr}
\toprule
Model & Default1 & Default2 & False & No & Right & True & Wrong & Yes & AVG \\
\midrule
Qwen-2.5 & 0.795 & 0.753 & 0.753 & 0.747 & 0.802 & 0.840 & 0.764 & 0.792 & 0.781 \\
Mistral-v0.3 & 0.646 & 0.177 & 0.681 & 0.733 & 0.726 & 0.819 & 0.715 & 0.681 & 0.647 \\
Llama-3 & 0.663 & 0.788 & 0.809 & 0.792 & 0.670 & 0.670 & 0.701 & 0.486 & 0.697 \\
Gemma & 0.076 & 0.198 & 0.198 & 0.319 & 0.306 & 0.212 & 0.323 & 0.347 & 0.247 \\
\bottomrule
\end{tabular}
}

\vspace{1em}

\subcaptionbox{Ambient Premises}[.9\linewidth]{
\begin{tabular}{lrrrrrrrrr}
\toprule
Model & Default1 & Default2 & False & No & Right & True & Wrong & Yes & AVG \\
\midrule
Qwen-2.5 & 0.654 & 0.654 & 0.706 & 0.700 & 0.712 & 0.711 & 0.707 & 0.700 & 0.693 \\
Mistral-v0.3 & 0.570 & 0.702 & 0.712 & 0.735 & 0.724 & 0.733 & 0.718 & 0.727 & 0.703 \\
Llama-3 & 0.524 & 0.729 & 0.672 & 0.631 & 0.673 & 0.672 & 0.632 & 0.551 & 0.636 \\
Gemma & 0.261 & 0.494 & 0.265 & 0.265 & 0.271 & 0.271 & 0.267 & 0.299 & 0.299 \\
\bottomrule
\end{tabular}
}
\caption{Accuracy Results for Different Ambiguity Sets}
\label{tab:ambiguity_results}
\end{table*}

We also provide a random sample of responses from Qwen-2 and Llama-3 using the disambiguation prompt in Figure~\ref{fig:prompts3} in Tables~\ref{tab:dis} and~\ref{tab:dis-2}. below. The full model responses will be available in our codebase.

\begin{table*}[h]
    \centering
    \renewcommand{\arraystretch}{1.2}
    \rowcolors{2}{gray!20}{} 
    \resizebox{\textwidth}{!}{
    \begin{tabular}{p{7cm}cp{10cm}}
        \hline
        \textbf{Text} & \textbf{Is Ambiguous} & \textbf{Disambiguations} \\
        \hline
        The man saw the woman with the telescope. & yes & ['The man used a telescope to see the woman, The man saw the woman who was holding a telescope'] \\

        The woman saw the man with the telescope. & yes & ['The woman used a telescope to see the man, The woman saw the man who was holding a telescope'] \\

        The guy saw the girl with the telescope.	&yes&	['The man used a telescope to see the woman, The man saw the woman who was using a telescope'] \\

        The man saw the woman with the microscope. & yes & ['The man used the microscope to see the woman, The man saw the woman who was using a microscope'] \\

        The man saw the woman with the monocular. & yes & ['The man used a monocular to see the woman, The man saw the woman who was holding a monocular'] \\

        The man saw the woman with the camera. & yes & ['The man saw the woman who was holding a camera, The man saw the woman using a camera to take a picture.'] \\

        The man saw the woman with the book. & yes & ['The man saw the woman holding a book, The man saw the woman who was a writer of a book'] \\

        The man saw the woman with the dress. & yes & ['The man saw the woman wearing the dress, The man saw the woman who was with the dress (possibly holding it or in possession of it).']\\

        The man saw the woman with the suitcase. & yes & ['The man saw a woman who was carrying a suitcase, The man saw a woman with a suitcase (near her)'] \\

        The mug saw the woman with the telescope. & yes & ['A mug witnessed a woman using a telescope, A person named Mug observed a woman with a telescope.'] \\

        The telescope saw the man with the woman. & yes & ['The telescope was used to observe a man and a woman together, The telescope had a man and a woman looking through it at the same time.'] \\

        The detective saw the woman with the telescope. & yes & ['The detective saw the woman using a telescope, The detective saw the woman near a telescope'] \\

        The teacher saw the woman with the telescope. & yes & ['The teacher used a telescope to see the woman, The teacher saw the woman who was carrying a telescope'] \\

        The mother saw the woman with the telescope. & yes & ['The mother used a telescope to see the woman, The mother saw the woman holding a telescope'] \\

        The father saw the woman with the telescope. & yes & ['The father saw a woman who was using a telescope, The father saw a woman with a telescope, meaning a woman who is associated with or has a connection to a telescope']\\
        
        The artist saw the woman with the telescope. & yes & ['The artist saw a woman who was using a telescope, The artist saw a woman who was looking at something through a telescope'] \\
        
        The explorer saw the woman with the telescope. & yes & ['The explorer saw a woman who was using a telescope, The explorer saw a woman with a telescope, i.e. the woman was in possession of the telescope'] \\
        
        The man saw the bridge with the telescope. & yes & ['The man saw the bridge through the telescope, The man saw the bridge that had a telescope on it'] \\
        
        The man saw the eagle with the telescope. & yes & ['The man saw the eagle through the telescope, The man saw the eagle that had a telescope on it'] \\
        
        The man saw the weather with the telescope. & yes & ['The man used a telescope to observe the weather, The man saw the weather conditions through the telescope'] \\
        
        The man saw the beer with the telescope. & yes & ['The man saw the beer using the telescope, The man looked at the beer with the aid of a telescope'] \\
        
        The man saw the bee with the telescope. & yes & ['The man saw the bee with his own eyes, The man used the telescope to see the bee'] \\

        Jenny saw Alex with the camera.	& yes &	[Jenny saw Alex holding a camera, Jenny saw Alex use a camera to take a picture'] \\

     The boy saw the girl with the glasses.	& yes &	['The boy saw the girl wearing glasses, The girl was accompanied by a boy who was wearing glasses'] \\

    The biologist saw the student with the telescope.	& yes &	['The biologist saw the student using the telescope, The biologist saw the student with (near or holding) a telescope'] \\

    \end{tabular}}
    \caption{Ambiguity predictions and their disambiguations according to Qwen-2.5}
    \label{tab:dis}
\end{table*}

\begin{table*}[h]
    \centering
    \renewcommand{\arraystretch}{1.2}
    \rowcolors{2}{gray!20}{}
    \resizebox{\textwidth}{!}{
    \begin{tabular}{p{7cm}cp{10cm}}
        \hline
        \textbf{Text} & \textbf{Is Ambiguous} & \textbf{Disambiguations} \\
        \hline
        The man saw the woman with the telescope. & yes & ['The man saw the woman while using the telescope, The man saw the woman who was looking through the telescope.'] \\

        The woman saw the man with the telescope. & yes & ['The woman saw the man who was using the telescope, The woman saw the man with the telescope (i.e., the man was in her presence, possibly holding the telescope)'] \\

        The guy saw the girl with the telescope.	&yes&	['The man saw the woman who was using the telescope, The guy saw the woman with a telescope, observing or accompanying her as she used it.'] \\

        The man saw the woman with the microscope. & yes & ['The man saw the woman who was using the microscope, The man saw the woman holding the microscope.'] \\

        The man saw the woman with the monocular. & yes & ['The man saw a woman who was using a monocular, The man saw a woman with a physical characteristic of having a monocular (e.g. one eye) or with a monocular as a possession.'] \\

        The man saw the woman with the camera. & yes & ['The man saw the woman who was holding the camera, The man saw the woman while the man was using the camera.'] \\

        The man saw the woman with the book. & yes & ['The man saw the woman while she was holding the book, The man saw the woman and noticed that she had the book.'] \\

        The man saw the woman with the dress. & yes & ['The man saw the woman who was wearing the dress, The man saw the woman with the person who was wearing the dress.']\\

        The man saw the woman with the suitcase. & yes & ['The man saw the woman who was carrying the suitcase, The man saw the woman and then saw the suitcase.'] \\

        The telescope saw the man with the woman. & yes & ['The telescope saw the man who was with the woman, The telescope saw the man while the woman was with it.'] \\
    
         The mug saw the woman with the telescope. & yes & ['The mug, a household item, is being used as a metaphor to describe something that saw the woman with the telescope, possibly implying that the mug has a unique perspective or insight. The mug, a household item, is literally being used to describe the object that saw the woman with the telescope, possibly implying that the mug has eyes or is being used as a tool to observe the woman with the telescope.'] \\

        The detective saw the woman with the telescope. & yes & ['The detective saw a woman who was using a telescope, The detective saw a woman who was being looked at with a telescope.'] \\

        The teacher saw the woman with the telescope. & yes & ['The teacher saw a woman who was using a telescope, The teacher saw a woman who was being looked at through a telescope.'] \\

        The mother saw the woman with the telescope. & yes & ['The mother saw the woman who was using the telescope, The mother saw the woman with the telescope, but the woman was not using it.'] \\

        The father saw the woman with the telescope. & yes & ['The father saw a woman who was using a telescope; The father saw a woman with a telescope, meaning a woman who is associated with or has a connection to a telescope.']\\
        
        The artist saw the woman with the telescope. & yes & ['The artist saw a woman who was using a telescope, The artist saw a woman who was looking at something through a telescope] \\
        
        The explorer saw the woman with the telescope. & yes & ['The explorer saw a woman who was using a telescope, The explorer saw a woman with a telescope, i.e. the woman was in possession of the telescope.'] \\
        
        The man saw the weather with the telescope. & yes & ['The man used a telescope to observe the weather, The man saw the weather conditions through the telescope.'] \\
        
        The man saw the beer with the telescope. & yes & ['The man saw the beer using the telescope, The man looked at the beer with the aid of a telescope.'] \\
        
        The man saw the bee with the telescope. & yes & ['The man saw the bee with his own eyes, The man used the telescope to see the bee.']\\

        Jenny saw Alex with the camera.	& yes &	['Jenny saw Alex with the camera, meaning Jenny used the camera to take a picture of Alex; Jenny saw Alex who was holding the camera.'] \\
    
        The boy saw the girl with the glasses.	& yes &	['The boy saw a girl who was wearing glasses, The boy saw a girl who was with someone wearing glasses.'] \\

        The biologist saw the student with the telescope.	& yes &	['The biologist was using a telescope to look at the student, The biologist saw a student who was using a telescope.'] \\

    \end{tabular}}
    \caption{Ambiguity predictions and their disambiguations according to Llama 3}
    \label{tab:dis-2}
\end{table*}

\subsection{Results Across Syntactic Roles}

\label{sec:appendix}
\begin{figure*}[ht!]
    \centering
    \includegraphics[]{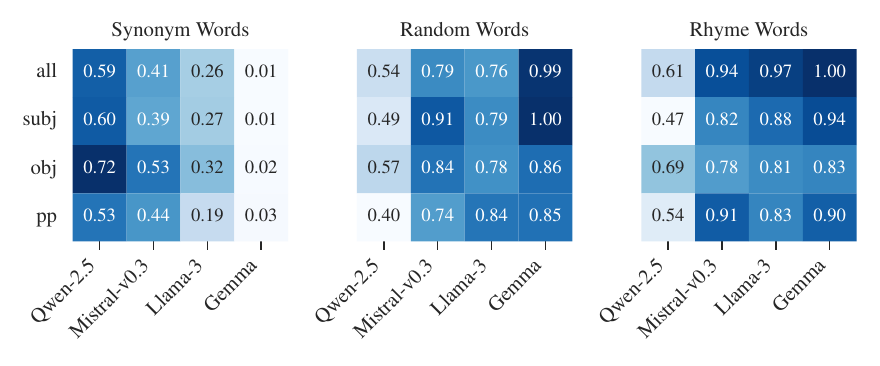}
    \caption{Results across models after zero-shot prompting, showing the error rate of the different synonym, random, and rhyme word modifications. Values closer to 1 indicate a higher error rate. We denote the substitution of the subject components of a sentence as \textit{subj}, the object components as \textit{obj}, and the prepositional phrases as \textit{pp}. Lastly, when we substitute more than two words we refer to this manipulation as \textit{all}.}
    \label{fig:heatmap-synt-roles}
\end{figure*}

While creating the dataset for syntactic ambiguity, we carefully altered the sentences according to the manipulations described in Table~\ref{tab:data-table} by changing specific sentence components. In particular, we modified the subject, object, the prepositional phrase (if any), and all components in the set of our original sentences. This division reflects the fact that ambiguity arises from a network of heterogeneous participants—agents, subjects, and objects—each with varying roles, reflexive awareness, and intentionality~\citep{amb}.

After prompting our models, we then examined whether there were any salient differences in error rate across syntactic roles. We report the results of this experiment in Figure~\ref{fig:heatmap-synt-roles}. Overall, we find that the results are relatively stable across syntactic roles and no significant patterns could be identified.

\section{Representational Analysis} \label{app:rep_analysis}

\begin{figure*}[ht!]
    \centering
    \begin{minipage}{0.24\textwidth}
        \centering
        \includegraphics[width=\linewidth]{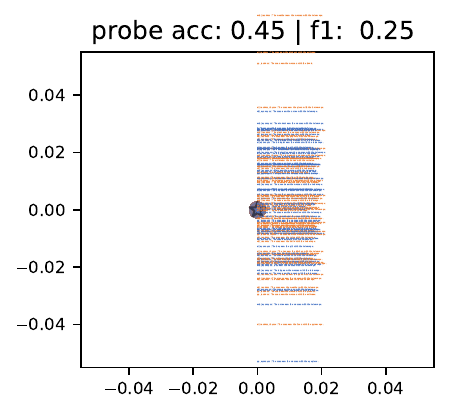}
    \end{minipage}
    \hfill
    \begin{minipage}{0.24\textwidth}
        \centering
        \includegraphics[width=\linewidth]{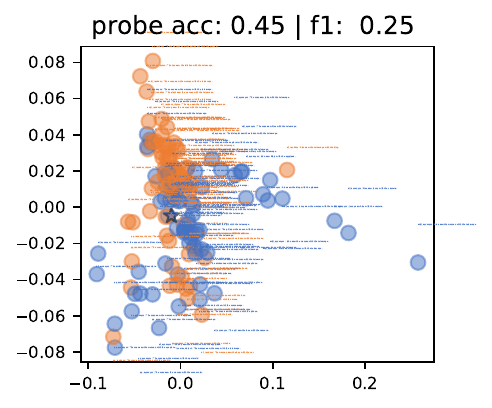}
    \end{minipage}
    \hfill
    \begin{minipage}{0.24\textwidth}
        \centering
        \includegraphics[width=\linewidth]{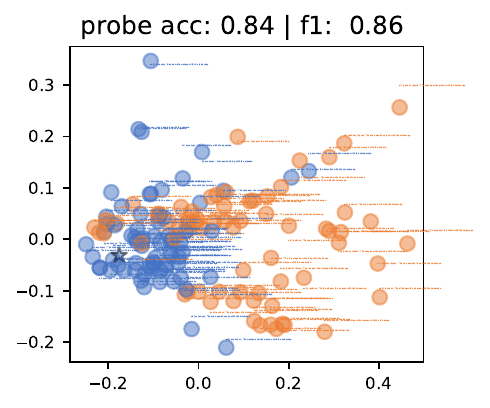}
    \end{minipage}
    \hfill
    \begin{minipage}{0.24\textwidth}
        \centering
        \includegraphics[width=\linewidth]{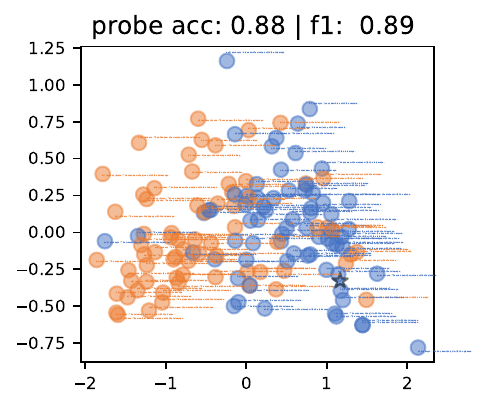}
    \end{minipage}

    \vspace{3mm} 

    \begin{minipage}{0.24\textwidth}
        \centering
        \includegraphics[width=\linewidth]{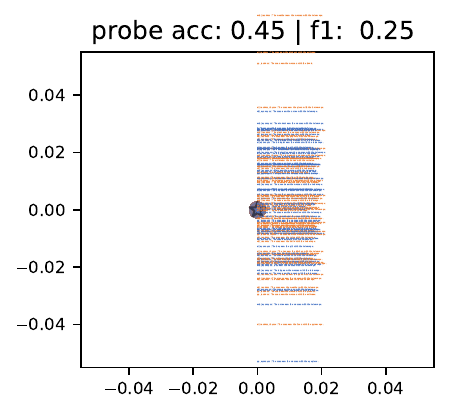}
    \end{minipage}
    \hfill
    \begin{minipage}{0.24\textwidth}
        \centering
        \includegraphics[width=\linewidth]{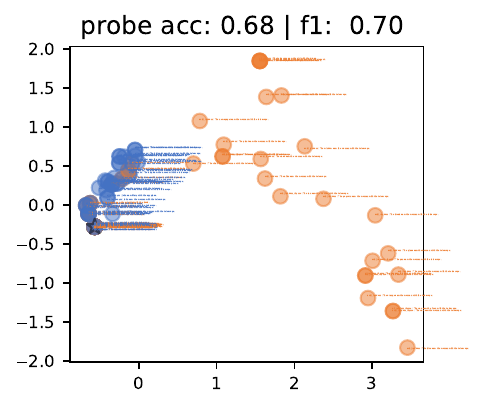}
    \end{minipage}
    \hfill
    \begin{minipage}{0.24\textwidth}
        \centering
        \includegraphics[width=\linewidth]{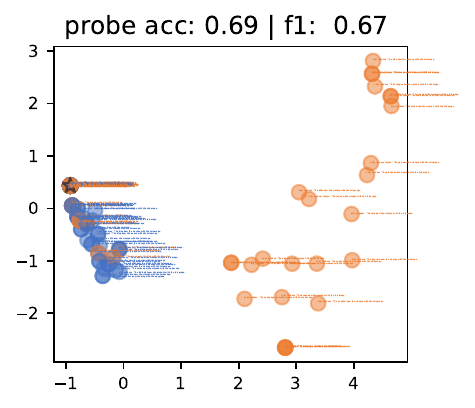}
    \end{minipage}
    \hfill
    \begin{minipage}{0.24\textwidth}
        \centering
        \includegraphics[width=\linewidth]{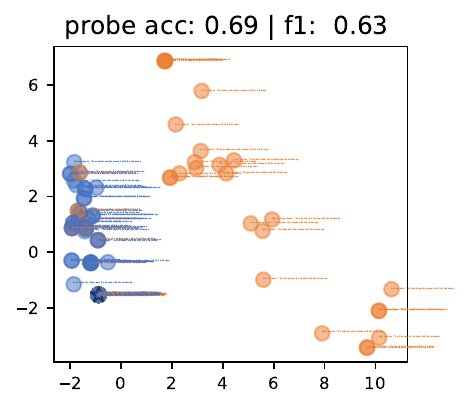}
    \end{minipage}

    \vspace{3mm} 

    \begin{minipage}{0.24\textwidth}
        \centering
        \includegraphics[width=\linewidth]{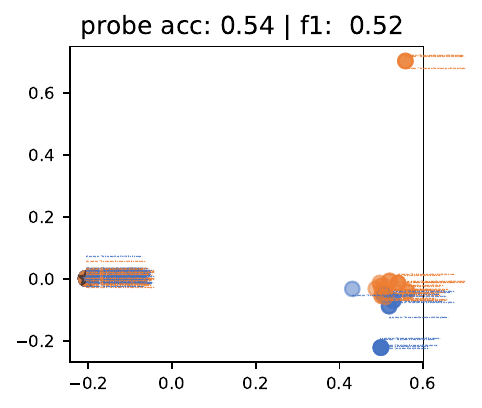}
    \end{minipage}
    \hfill
    \begin{minipage}{0.24\textwidth}
        \centering
        \includegraphics[width=\linewidth]{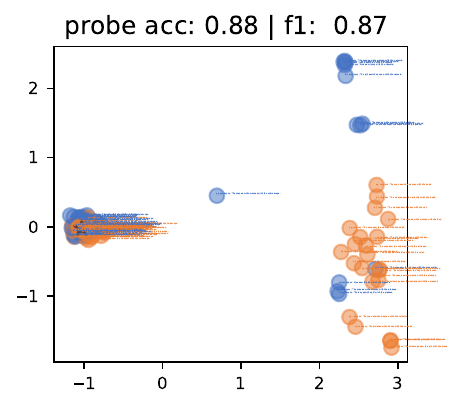}
    \end{minipage}
    \hfill
    \begin{minipage}{0.24\textwidth}
        \centering
        \includegraphics[width=\linewidth]{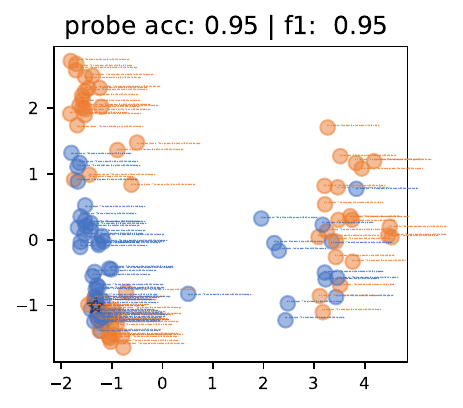}
    \end{minipage}
    \hfill
    \begin{minipage}{0.24\textwidth}
        \centering
        \includegraphics[width=\linewidth]{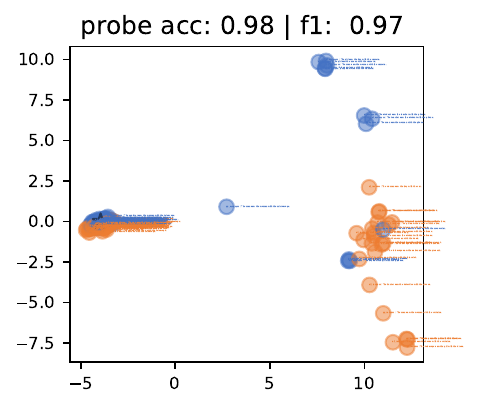}
    \end{minipage}

        \vspace{3mm} 

    \begin{minipage}{0.24\textwidth}
        \centering
        \includegraphics[width=\linewidth]{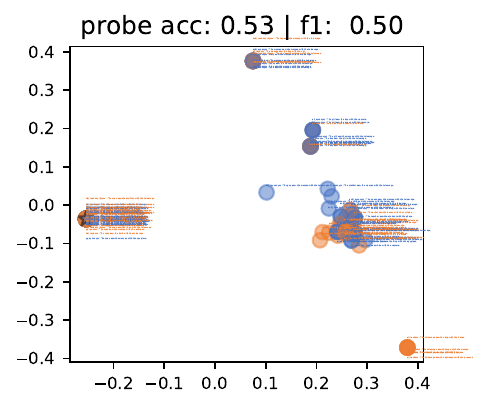}
        \caption*{Layer 0}
    \end{minipage}
    \hfill
    \begin{minipage}{0.24\textwidth}
        \centering
        \includegraphics[width=\linewidth]{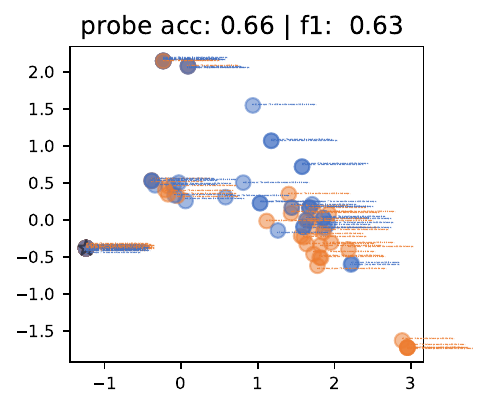}
        \caption*{Layer 6}
    \end{minipage}
    \hfill
    \begin{minipage}{0.24\textwidth}
        \centering
        \includegraphics[width=\linewidth]{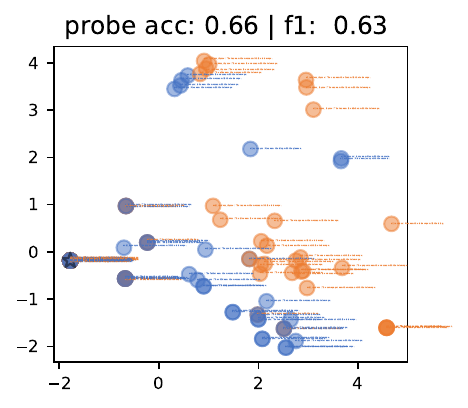}
        \caption*{Layer 12}
    \end{minipage}
    \hfill
    \begin{minipage}{0.24\textwidth}
        \centering
        \includegraphics[width=\linewidth]{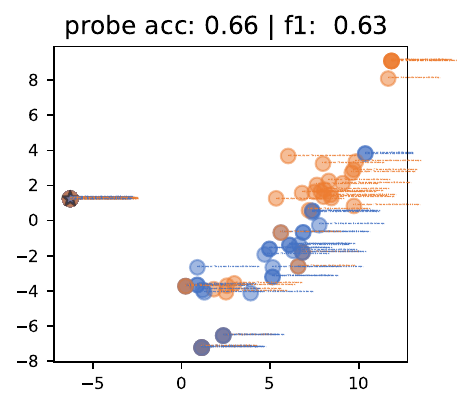}
        \caption*{Layer 24}
    \end{minipage}

    \caption{PCA projections extracted for different token roles (eos, verb, last noun, first noun) across layers for Llama 3. Blue/orange indicate ambiguous/not ambiguous sentences.}
    \label{fig:all-figures}
\end{figure*}

\begin{figure*}[ht!]
    \centering
    \begin{minipage}{0.24\textwidth}
        \centering
        \includegraphics[width=\linewidth]{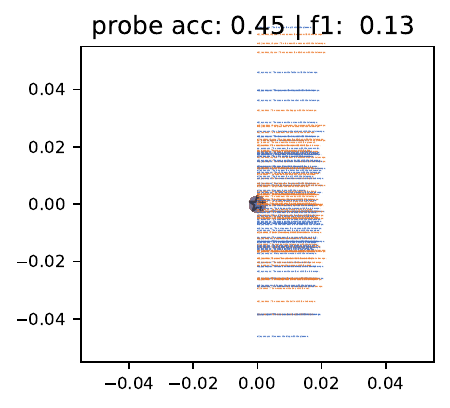}
    \end{minipage}
    \hfill
    \begin{minipage}{0.24\textwidth}
        \centering
        \includegraphics[width=\linewidth]{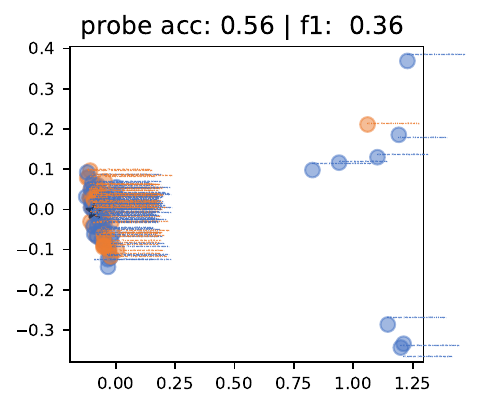}
    \end{minipage}
    \hfill
    \begin{minipage}{0.24\textwidth}
        \centering
        \includegraphics[width=\linewidth]{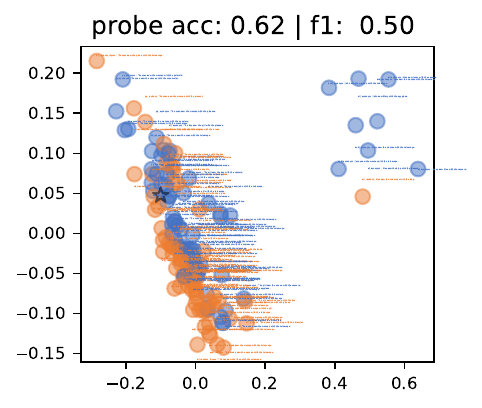}
    \end{minipage}
    \hfill
    \begin{minipage}{0.24\textwidth}
        \centering
        \includegraphics[width=\linewidth]{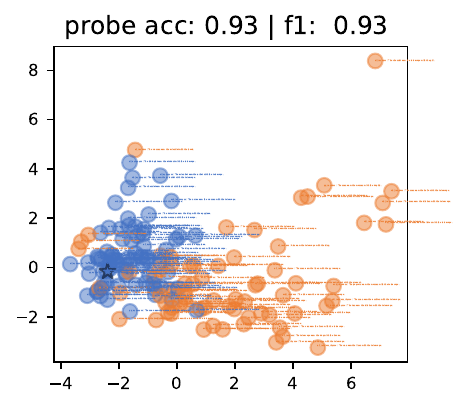}
        \label{fig:fig20}
    \end{minipage}

    \vspace{3mm} 

    \begin{minipage}{0.24\textwidth}
        \centering
        \includegraphics[width=\linewidth]{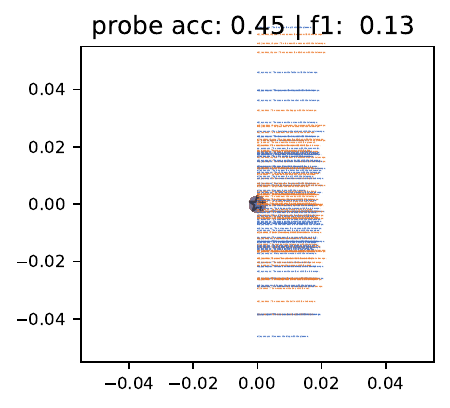}
        \label{fig:fig21}
    \end{minipage}
    \hfill
    \begin{minipage}{0.24\textwidth}
        \centering
        \includegraphics[width=\linewidth]{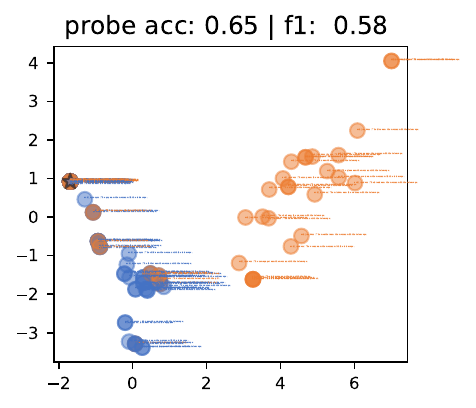}
        \label{fig:fig22}
    \end{minipage}
    \hfill
    \begin{minipage}{0.24\textwidth}
        \centering
        \includegraphics[width=\linewidth]{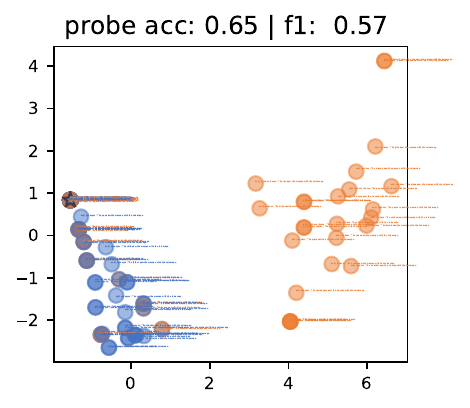}
        \label{fig:fig23}
    \end{minipage}
    \hfill
    \begin{minipage}{0.24\textwidth}
        \centering
        \includegraphics[width=\linewidth]{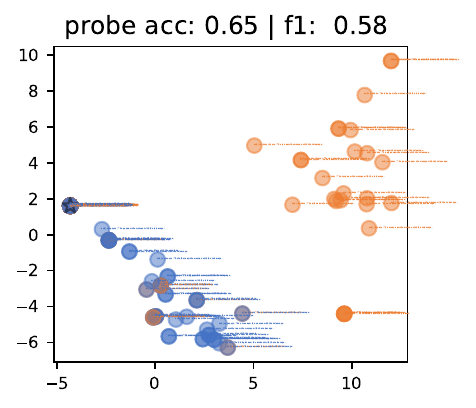}
        \label{fig:fig24}
    \end{minipage}

    \vspace{3mm} 

    \begin{minipage}{0.24\textwidth}
        \centering
        \includegraphics[width=\linewidth]{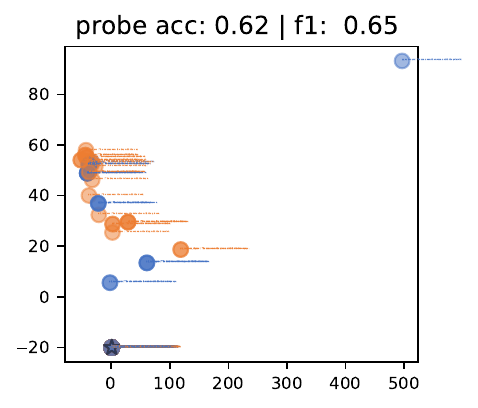}
    \end{minipage}
    \hfill
    \begin{minipage}{0.24\textwidth}
        \centering
        \includegraphics[width=\linewidth]{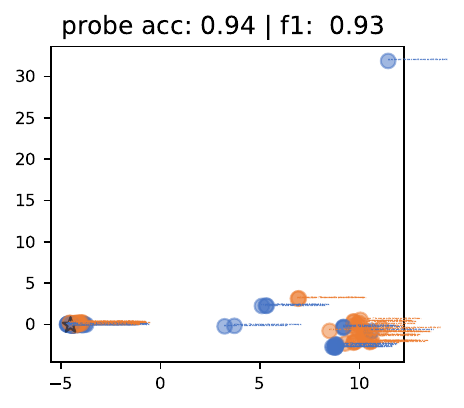}
    \end{minipage}
    \hfill
    \begin{minipage}{0.24\textwidth}
        \centering
        \includegraphics[width=\linewidth]{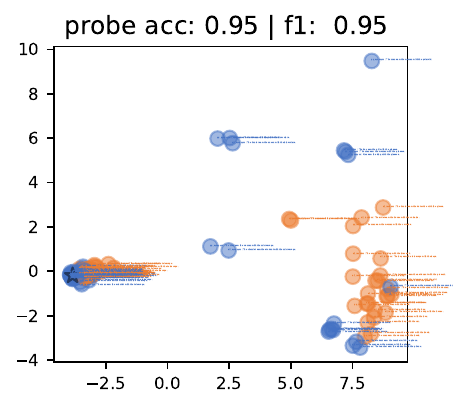}
        \label{fig:fig11}
    \end{minipage}
    \hfill
    \begin{minipage}{0.24\textwidth}
        \centering
        \includegraphics[width=\linewidth]{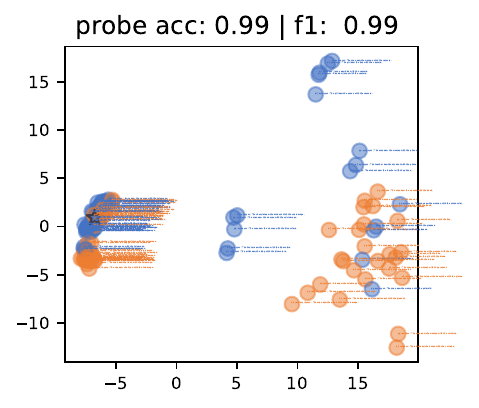}
        \label{fig:fig12}
    \end{minipage}
        \vspace{3mm} 

    \begin{minipage}{0.24\textwidth}
        \centering
        \includegraphics[width=\linewidth]{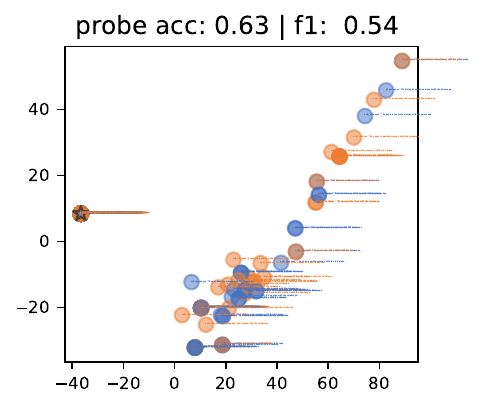}
        \caption*{Layer 0}
        \label{fig:fig13}
    \end{minipage}
    \hfill
    \begin{minipage}{0.24\textwidth}
        \centering
        \includegraphics[width=\linewidth]{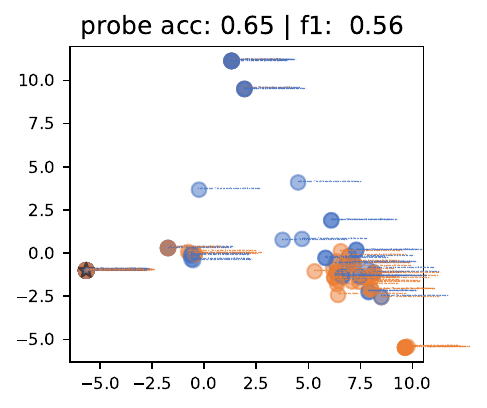}
        \caption*{Layer 6}
        \label{fig:fig14}
    \end{minipage}
    \hfill
    \begin{minipage}{0.24\textwidth}
        \centering
        \includegraphics[width=\linewidth]{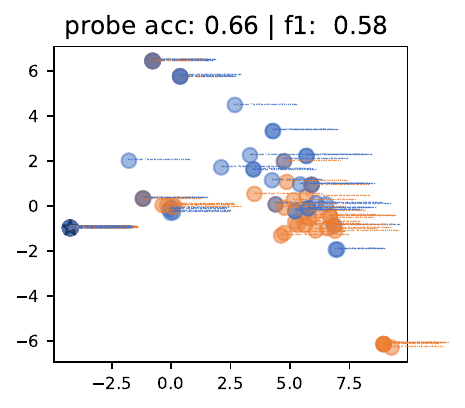}
        \caption*{Layer 12}
        \label{fig:fig15}
    \end{minipage}
    \hfill
    \begin{minipage}{0.24\textwidth}
        \centering
        \includegraphics[width=\linewidth]{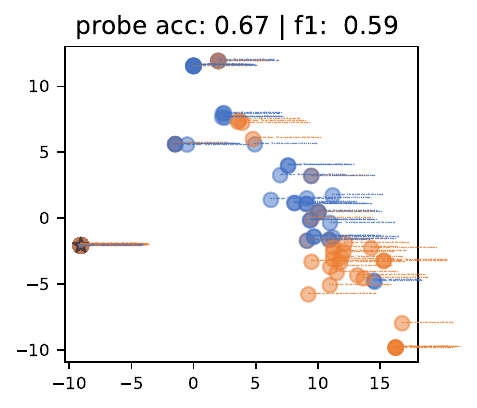}
        \caption*{Layer 24}
        \label{fig:fig16}
    \end{minipage}
    
    \caption{PCA projections extracted for different token roles (eos, verb, last noun, first noun) across layers for Gemma. Blue/orange indicate ambiguous/not ambiguous sentences.}   
    \label{fig:all-figures-gemma}
\end{figure*}

We also provide further insights from the representational analysis across more layers. We present examples from the initial (layer 0), early (layer 6),  middle (layer 12), and late layers (layer 24) across all token roles.
We show PCA projections for Llama-3 in Figure~\ref{fig:all-figures} and Gemma in Figure~\ref{fig:all-figures-gemma}  across layers and token types. We extract representations of the final end of prompt sentence representation (eos), the first verb, first noun, and last noun. Across models and token types, representations extracted at the first layer, which are not yet contextualized via the encoder module, do not allow for decoding ambiguity, and no meaningful clustering structure has emerged yet. At early layer 6, the last noun already achieves remarkably high probe accuracies in both Llama-3 ($0.83$) and Gemma ($0.93$), which consistently increase towards later layers. Similarly, but at an overall lower level, the first verb and first noun, also achieve higher probe accuracies of $0.6-0.69$,  but plateau from middle layers on for both models. Interestingly, the eos probe accuracies remain low in early layers at $0.45$ (Llama-3) and $0.55$ (Gemma), but increase from then on and achieve high scores in late layers of around $0.9$. We furthermore observe a clustering of samples around the original sentence examples used to build adversaries and variations contained in our AmbAdv dataset, e.g., for the first verb, this appears across all layers in both models, hinting at the model's limited ability to accurately disambiguate sentence meaning from the verb alone. Our results further hint at a special role of the last noun, as the model appears to distinguish between ambiguous and non-ambiguous instances through the last noun representation. 
Importantly, several token types outperform the behavioural model evaluation, while our representational probe analysis of the eos token highlights that relevant information for accurate inferences is available, but can not be reliably used by the model for generating correct answers.
In conclusion, our analysis highlights the evolving role of token representations across layers, with a particular emphasis on the last noun as a central factor in disambiguating sentence meaning.

\end{document}